\documentclass[conference]{IEEEtran}
\IEEEoverridecommandlockouts
\usepackage{cite}
\usepackage{amsmath,amssymb,amsfonts}
\usepackage{algorithm}
\usepackage{algpseudocode}
\usepackage{graphicx}
\usepackage{textcomp}
\usepackage{xcolor}
\usepackage{hyperref}
\usepackage{threeparttable}
\usepackage{array}
\usepackage{booktabs}
\usepackage{multirow}
\usepackage{makecell}
\usepackage{subcaption}
\usepackage{orcidlink}
\usepackage{authblk}

\hypersetup{
    colorlinks=true,
    urlcolor=magenta,      
}

\def\BibTeX{{\rm B\kern-.05em{\sc i\kern-.025em b}\kern-.08em
    T\kern-.1667em\lower.7ex\hbox{E}\kern-.125emX}}
\begin{document}

\title{GAT-RWOS: Graph Attention-Guided Random Walk Oversampling for Imbalanced Data Classification}

\author[1,2]{Zahiriddin Rustamov \orcidlink{0000-0003-4977-1781}}
\author[1]{Abderrahmane Lakas}
\author[1,*]{Nazar Zaki}

\affil[1]{College of Information Technology, United Arab Emirates University, Al Ain, UAE}
\affil[2]{Department of Computer Science, KU Leuven\authorcr{\tt nzaki@uaeu.ac.ae}}

\maketitle

\begin{abstract}
    Class imbalance poses a significant challenge in machine learning (ML), often leading to biased models favouring the majority class. In this paper, we propose GAT-RWOS, a novel graph-based oversampling method that combines the strengths of Graph Attention Networks (GATs) and random walk-based oversampling. GAT-RWOS leverages the attention mechanism of GATs to guide the random walk process, focusing on the most informative neighbourhoods for each minority node. By performing attention-guided random walks and interpolating features along the traversed paths, GAT-RWOS generates synthetic minority samples that expand class boundaries while preserving the original data distribution. Extensive experiments on a diverse set of imbalanced datasets demonstrate the effectiveness of GAT-RWOS in improving classification performance, outperforming state-of-the-art oversampling techniques. The proposed method has the potential to significantly improve the performance of ML models on imbalanced datasets and contribute to the development of more reliable classification systems. Code is available at \url{https://github.com/zahiriddin-rustamov/gat-rwos}.
\end{abstract}


\section{Introduction}\label{sec:introduction}

Class imbalance is a critical challenge in machine learning (ML), where some classes are significantly underrepresented, can skew models toward the majority class and neglect critical minority classes, particularly in domains like medical diagnosis and fraud detection \cite{Joloudari2023}.
Various techniques have been proposed to address the class imbalance problem. These include data-level methods (e.g., oversampling minority, undersampling majority classes), algorithm-level methods (e.g., cost-sensitive learning, ensemble methods), and hybrid approaches. Among these, oversampling techniques, which generate synthetic samples to balance the class distribution, have gained attention due to their effectiveness \cite{Kovacs2019}.

However, oversampling techniques like the Synthetic Minority Over-sampling Technique (SMOTE) may overgeneralize near decision boundaries and are sensitive to noise and outliers. Recent graph-based methods \cite{Zhao2021, Wu2021, Liu2022} leverage data structure to generate better synthetic samples but face challenges in mapping augmented graphs back to the original feature space, limiting their practicality.

To overcome these limitations, we propose GAT-RWOS, a novel graph-based oversampling method combining Graph Attention Networks (GATs) and attention-guided random walks. Our approach uses GAT's attention mechanism to guide random walks through informative neighbourhoods of minority nodes, generating synthetic samples that expand class boundaries while preserving the data distribution.
The main contributions of this paper are:
\begin{itemize}
    \item We propose GAT-RWOS, a novel graph-based oversampling method that leverages the attention mechanism of GATs to guide the random walk process and generate high-quality synthetic minority samples.
    \item We demonstrate GAT-RWOS's effectiveness in improving classification on imbalanced binary classification tasks, outperforming traditional methods like SMOTE.
\end{itemize}
The rest of the paper is organized as follows: Section \ref{sec:related_works} reviews related work; Section \ref{sec:proposed_method} details the proposed method; Section \ref{sec:experimental_setup} describes the experimental setup; Section \ref{sec:results} discusses the results; and Section \ref{sec:conclusion} concludes the study.

\section{Related Works}\label{sec:related_works}

Traditional synthetic sample generation methods like SMOTE \cite{Chawla2002} interpolate between minority class samples but often overgeneralize near decision boundaries and perform poorly in high-dimensional spaces.
Kovács \cite{Kovacs2019} evaluated 85 oversampling variants, identifying top methods such as Polynom-Fit-SMOTE \cite{Gazzah2008}, which uses polynomial fitting to fill the minority subspace better; ProWSyn \cite{Barua2013}, which weights minority samples based on proximity to the majority class; and SMOTE-IPF \cite{Saez2015}, combining SMOTE with filtering to remove noisy and borderline examples.
Lee et al. \cite{Lee2015} introduced an oversampling technique with rejection to avoid overfitting and underfitting, while SMOBD \cite{Cao2011} uses distribution and density information to reduce noise influence and preserve data distribution.
Despite improvements over SMOTE, these advanced oversamplers do not consistently outperform across datasets \cite{Kovacs2019}, indicating the need for techniques better suited to specific dataset characteristics, high dimensionality, noise, and complex distributions.

Random walk-based methods generate synthetic samples via random walks to expand class boundaries while preserving data distribution. Zhang and Li \cite{Zhang2014} introduced Random Walk Oversampling (RWOS), generating samples based on minority class mean and variance. Roshanfekr et al. \cite{Roshanfekr2020} proposed UGRWOS, constructing local graphs to maintain proximity information and applying RWOS to selected minority samples. Sadhukhan et al. \cite{Sadhukhan2022} presented RWMaU, using random walks to identify and undersample majority class samples in overlapping regions. These methods better capture minority class structures and expand boundaries but need improvement in adaptively guiding random walks based on local data characteristics and leveraging information from both classes.

Graph-based methods tackle class imbalance by constructing graph representations. GraphSMOTE \cite{Zhao2021} extends SMOTE to graph data, generating synthetic minority nodes in embedding space and training an edge generator. GraphMixup \cite{Wu2021} performs semantic-level mixup, incorporating self-supervised and reinforcement learning techniques. ESA-GCN \cite{Zhang2023} introduces hybrid sampling with attention-based edge generation, while GATSMOTE \cite{Liu2022} uses graph attention and homophily principles for brain networks. Despite their potential, these methods face challenges in translating the augmented graph back to the original feature space, complicating integration with standard ML pipelines.

While advancements have been made, effectively leveraging structural information to guide oversampling remains challenging. Traditional random walk-based methods ignore the importance of different neighborhoods, and graph-based methods struggle to map augmented graphs back to feature space.
To address this, we propose GAT-RWOS, combining the strengths of GATs \cite{Velickovic2017} and random walk-based oversampling. GAT-RWOS leverages the attention mechanism of GATs to guide random walks, focusing on the most informative neighbourhoods for each minority node. By learning attention weights that capture neighbour importance, GAT-RWOS adaptively steers random walks toward regions likely to generate valuable synthetic samples.

The novelty of GAT-RWOS lies in its ability to adaptively guide the oversampling process based on learned attention weights, ensuring generated samples are representative and effectively expand minority class boundaries.
Furthermore, operating directly in the feature space allows GAT-RWOS to integrate easily into existing ML pipelines without additional post-processing.

\section{Proposed Method}\label{sec:proposed_method}

\subsection{Graph Attention Network (GAT)}

GATs are neural architectures designed for graph-structured data, employing attention mechanisms to assign weights to neighbouring nodes and focus on the most relevant information.
Given a graph $\mathcal{G} = (\mathcal{V}, \mathcal{E})$ with node features $\vec{h}_i \in \mathbb{R}^F$, the attention coefficients $\alpha_{ij}$ between nodes $i$ and $j$ are computed as:
\[
    \alpha_{ij} = \frac{\exp\left(\text{LeakyReLU}\left(a^\top [W\vec{h}_i \, \Vert \, W\vec{h}_j]\right)\right)}{\sum_{k \in \mathcal{N}_i} \exp\left(\text{LeakyReLU}\left(a^\top [W\vec{h}_i \, \Vert \, W\vec{h}_k]\right)\right)}
\]
where $W \in \mathbb{R}^{F' \times F}$ is a learnable weight matrix, $a$ is the attention mechanism, $\mathcal{N}_i$ denotes the neighbors of node $i$, and $\text{LeakyReLU}$ is the activation function.

The output features for each node are obtained by aggregating features across the node's neighbours, scaled by the computed attention coefficients, and then passed through a non-linear activation function:
\[
    \vec{h}_i' = \Big\Vert_{k=1}^K \sigma\left(\sum_{j \in \mathcal{N}_i} \alpha_{ij}^k W^k \vec{h}_j\right)
\]
where $\Vert$ denotes concatenation, $K$ is the number of attention heads, $\alpha_{ij}^k$ are the attention coefficients for head $k$, $W^k$ are head-specific transformation matrices, and $\sigma$ is a non-linear activation function.

We employ a multi-layer GAT to learn node embeddings that capture both structural and attribute information, with the final layer producing class probabilities. To address class imbalance, we incorporate focal loss \cite{Lin2017} and introduce class-specific attention weights. The focal loss is defined as:
\[
    \mathcal{L}_{\text{focal}} = -\sum_{i \in \mathcal{V}_{\text{train}}} (1 - p_i)^\gamma \log(p_i)
\]
where $p_i$ is the predicted probability of the true class for node $i$, $\gamma$ adjusts the down-weighting of easy examples, and $\mathcal{V}_{\text{train}}$ are the training nodes. Class-specific attention weights are introduced as learnable parameters to further tailor the attention mechanism toward minority classes.

\subsection{Biased Random Walk}

Our biased random walk utilizes the GAT attention matrix to guide the walk process, focusing on informative neighbourhoods for each minority node. Inspired by Node2Vec \cite{Grover2016}, it introduces a flexible procedure to explore diverse neighbourhoods.
Let \( M_m \in \mathbb{R}^{N_m \times N_m} \) denote the minority attention matrix, where \( N_m \) is the number of minority nodes, and \( M_m[i, j] \) represents the attention weight between nodes \( i \) and \( j \). Starting from a minority node \( v_s \), the walk generates a path \( P \) of length \( n_w \), traversing the graph based on attention weights and parameters \( p \) (return probability) and \( q \) (in-out probability).
The transition probabilities \( \pi \) for each neighbour \( v \) are calculated as:
\[
    \pi[v] = \begin{cases}
        {M_m[v_{\text{curr}}, v]}/{p}, & \text{if } v = v_{\text{prev}} \\
        {M_m[v_{\text{curr}}, v]}/{q}, & \text{if } v \notin P \\
        M_m[v_{\text{curr}}, v], & \text{otherwise}
    \end{cases}
\]
Here, \( v_{\text{prev}} \) is the previous node in \( P \); \( p \) and \( q \) control the likelihood of revisiting nodes or exploring new ones. The probabilities are normalized:
\[
    \pi[v] = \frac{\pi[v]}{\sum_{u \in N} \pi[u]}
\]
where \( N \) is the set of neighbors of \( v_{\text{curr}} \). Next, \( v_{\text{next}} \) is sampled from \( N \) according to \( \pi \), appended to \( P \), and the process repeats until \( P \) reaches length \( n_w \) or no neighbours remain. The procedure is detailed in Algorithm \ref{alg:biased_walk}.

\begin{algorithm}[!t]
\caption{Biased Random Walk (\textsc{BiasedWalk})}
\label{alg:biased_walk}
\footnotesize
\begin{algorithmic}[1]
\Require $v_s$: Starting node, $M_m$: Minority attention matrix, $n_w$: Number of walk steps, $p$: Return probability, $q$: In-out probability
\Ensure $P$: Random walk path
\State $P \gets [v_s]$
\State $v_\text{curr} \gets v_s$
\For{$i \gets 1$ to $n_w$}
    \State $N \gets$ neighbors of $v_\text{curr}$ in $M_m$
    \If{$|N| = 0$}
        \State \textbf{break}
    \EndIf
    \State $v_{\text{prev}} \gets P[-2]$ if $|P| > 1$ else None
    \State Initialize transition probabilities $\pi$
    \For{each $v \in N$}
        \State $\pi[v] \gets M_m[v_\text{curr}, v]$
        \If{$v = v_{\text{prev}}$}
            \State $\pi[v] \gets \pi[v] / p$
        \ElsIf{$v \notin P$}
            \State $\pi[v] \gets \pi[v] / q$
        \EndIf
    \EndFor
    \State Normalize $\pi$
    \State $v_\text{next} \gets$ sample a neighbor from $N$ according to $\pi$
    \State $P \gets P + [v_\text{next}]$
    \State $v_\text{curr} \gets v_\text{next}$
\EndFor
\State \Return $P$
\end{algorithmic}
\end{algorithm}

\subsection{Attention-Guided Interpolation}

The attention-guided interpolation generates synthetic minority samples by interpolating features along paths from biased random walks, guided by attention weights from the GAT that capture the importance of node relationships.
Let $P = [v_1, v_2, \ldots, v_n]$ be a path of minority nodes with feature vectors $\mathbf{x}_1, \mathbf{x}_2, \ldots, \mathbf{x}_n$. The attention weights $\alpha_{ij}$ between nodes $v_i$ and $v_j$ are derived from the minority attention matrix $M_m$. The adaptive interpolation weights $\lambda_{ij}$ are computed as:
\[
    \lambda_{ij} = \frac{\alpha_{ij}}{\sum_{k=1}^{n-1} \alpha_{k,k+1}}
\]
To emphasize significant attention values, these weights are adjusted using an exponential function:
\[
    \lambda_{ij}' = \frac{\exp(\lambda_{ij})}{\sum_{k=1}^{n-1} \exp(\lambda_{k,k+1})}
\]
The synthetic feature vector $\mathbf{x}_{\text{new}}$ is obtained via weighted summation:
\[
    \mathbf{x}_{\text{new}} = \sum_{i=1}^{n-1} \lambda_{i,i+1}' \cdot \text{Interpolate}(\mathbf{x}_i, \mathbf{x}_{i+1}, \lambda_{i,i+1}')
\]
The $\text{Interpolate}$ function handles both categorical and continuous features:
\[
    \mathbf{x}_{\text{new}}[k] = \begin{cases}
        \text{BestCategory}(\mathbf{x}_i[k], \mathbf{x}_{i+1}[k]), & \text{if } k \in \mathcal{C} \\
        \lambda_{i,i+1}' \mathbf{x}_i[k] + (1 - \lambda_{i,i+1}') \mathbf{x}_{i+1}[k], & \text{if } k \in \mathcal{T}
    \end{cases}
\]
where $\mathcal{C}$ and $\mathcal{T}$ are indices of categorical and continuous features, respectively, and $\text{BestCategory}(\cdot)$ selects the most similar category. To introduce variability and avoid overfitting, a small random perturbation $\epsilon$ is added to $\lambda_{i,i+1}'$, bounded by $\lambda_{\text{min}}$ and $\lambda_{\text{max}}$:
\[
    \lambda_{i,i+1}' = \min(\lambda_{\text{max}}, \max(\lambda_{\text{min}}, \lambda_{i,i+1}' + \text{Uniform}(-\epsilon, \epsilon)))
\]
This attention-guided interpolation process is summarized in Algorithm~\ref{alg:att_interp}.

\begin{algorithm}[!t]
\caption{Attention-Guided Interpolation (\textsc{AttInterp})}
\label{alg:att_interp}
\footnotesize
\begin{algorithmic}[1]
\Require $\mathbf{x}_i, \mathbf{x}_j$: Feature vectors of nodes $v_i$ and $v_j$, $\alpha_{ij}$: Attention weight between $v_i$ and $v_j$, $\mathcal{C}$: Set of categorical feature indices, $\mathcal{T}$: Set of continuous feature indices, $\lambda_{\text{min}}$: Minimum alpha, $\lambda_{\text{max}}$: Maximum alpha, $\epsilon$: Variability
\Ensure $\mathbf{x}_{\text{new}}$: Interpolated feature vector
\State $\lambda \gets \lambda_{\text{min}} + (\lambda_{\text{max}} - \lambda_{\text{min}}) \cdot \alpha_{ij}$
\State $\lambda \gets \lambda + \text{uniform}(-\epsilon, \epsilon)$
\State $\lambda \gets \max(\lambda_{\text{min}}, \min(\lambda, \lambda_{\text{max}}))$
\State $\mathbf{x}_{\text{new}} \gets \mathbf{0}$
\For{$k \gets 1$ to $\text{len}(\mathbf{x}_i)$}
    \If{$k \in \mathcal{C}$}
        \State Calculate average similarity per category for feature $k$ and select category with closest similarity
        \State $\mathbf{x}_{\text{new}}[k] \gets$ selected category
    \Else
        \State $\mathbf{x}_{\text{new}}[k] \gets \lambda \cdot \mathbf{x}_i[k] + (1 - \lambda) \cdot \mathbf{x}_j[k]$
    \EndIf
\EndFor
\State \Return $\mathbf{x}_{\text{new}}$
\end{algorithmic}
\end{algorithm}

\begin{figure*}[!ht]
    \centering
    \includegraphics[width=0.7\linewidth]{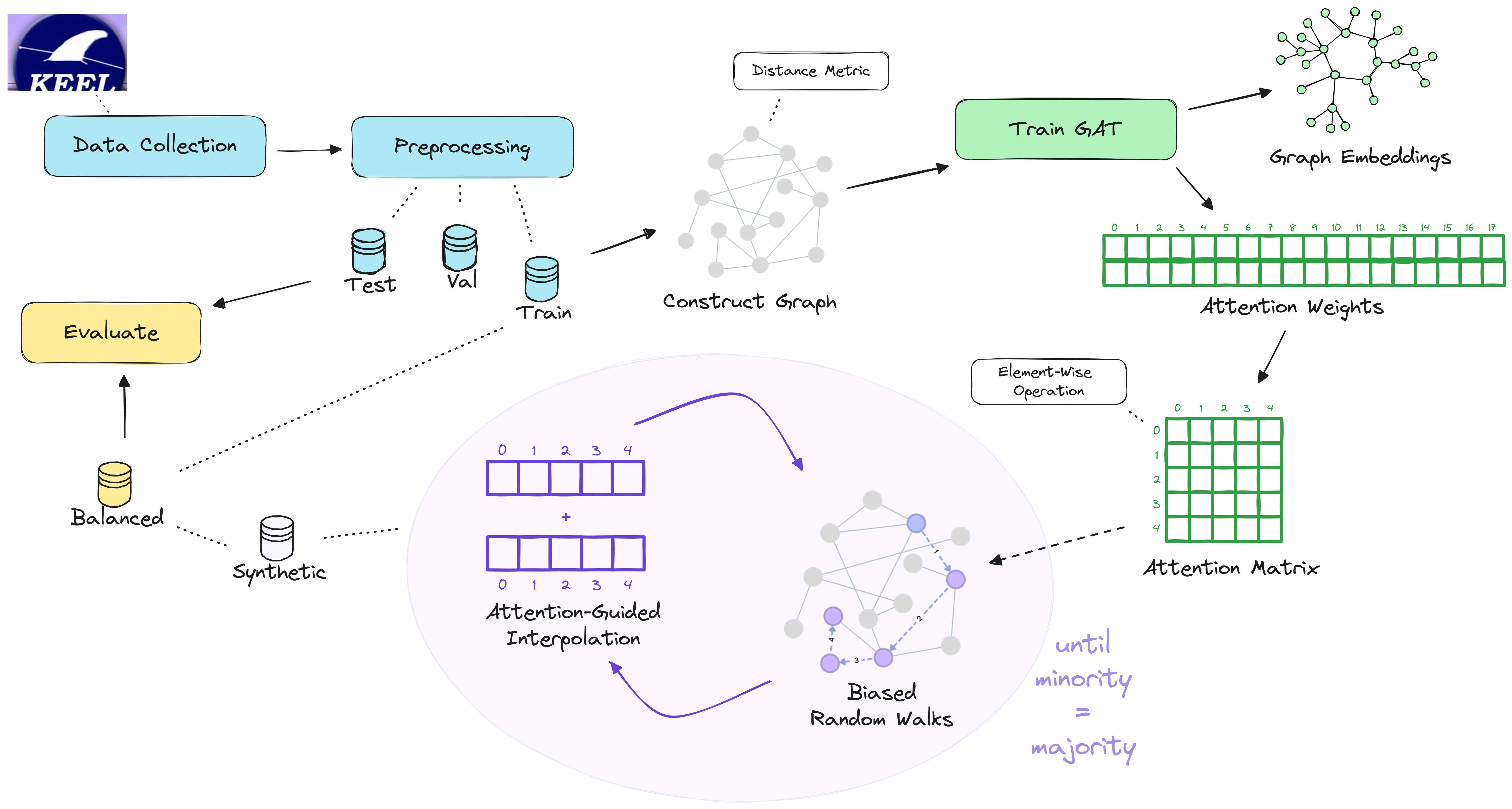}
    \caption{Methodology Overview of This Study.}
    \label{fig:methodology}
\end{figure*}

\subsection{GAT-RWOS}

The GAT-RWOS procedure (Algorithm \ref{alg:gat_rwos}) begins by training a GAT on graph $G$ (with feature matrix $\mathbf{X}$ and labels $\mathbf{y}$) to obtain attention weights $\mathbf{A}$. The attention weights from all $K$ heads are aggregated into a unified matrix $\mathbf{M}$:
\[
    \mathbf{M} = \text{Aggregate}(\mathbf{A}^{(1)}, \ldots, \mathbf{A}^{(K)})
\]
where $\mathbf{A}^{(k)}$ is the attention matrix from the $k$-th head. Minority nodes $V_m$ are identified, and a minority-specific attention matrix $\mathbf{M}_m$ is formed. Strong minority connections $E_s$ are determined by thresholding $\mathbf{M}_m$ with $\alpha$:
\[
    E_s = \{(v_i, v_j) \mid v_i, v_j \in V_m, \mathbf{M}_m[i, j] > \alpha\}
\]
The required number of synthetic samples $n_s$ is calculated based on the desired oversampling ratio.

To generate $n_s$ synthetic samples, the algorithm iteratively selects strongly connected minority node pairs $(v_i, v_j)$ from $E_s$. From $v_i$, a biased random walk of length $n_w$ is conducted using $\textsc{BiasedWalk}$ with $\mathbf{M}_m$ and parameters $p$ and $q$. If the resulting path $P$ contains at least two nodes, attention-guided adaptive interpolation weights are computed. Consecutive nodes $(v_a, v_b)$ along $P$ are interpolated via $\textsc{AttInterp}$ to generate new feature vectors $\mathbf{x}_\text{new}$, which are added to the synthetic sample set $\mathbf{X}_\text{syn}$. This interpolation is repeated $n_i$ times per path to enhance diversity. The process continues until $n_s$ synthetic samples are generated, which are then merged with the original training set to achieve balance.

\section{Experimental Setup}\label{sec:experimental_setup}

Our methodology, as shown in Figure \ref{fig:methodology}, begins with collecting and preprocessing imbalanced datasets to construct graphs for GAT training. After training, we extract graph embeddings and attention weights, combining them into an attention matrix via element-wise operations (e.g., mean). A sample generation algorithm performs biased random walks until the minority class size matches the majority class. The generated synthetic samples are added to the training set, creating a balanced dataset for evaluation.

\subsection{Datasets}

We selected binary classification datasets from the KEEL repository \cite{keel}, known for class imbalance issues. Datasets, where ML classifiers achieved balanced accuracy $\geq 0.8$, were excluded, as they either had mild imbalance or were too simplistic. We focused on challenging datasets with significant class imbalances. Table \ref{table:datasets} details the selected datasets, listing total instances ($N$), positive instances ($N_+$), features ($f$), and imbalance ratio ($IR = N_-/N_+$).

\begin{table*}[!ht]
\caption{Overview of Datasets Used in This Study.}\label{table:datasets}
\scriptsize
\centering
\begin{threeparttable}
\begin{tabular*}{0.85\textwidth}{@{\extracolsep{\fill}} 
    >{\raggedright\arraybackslash}p{0.21\linewidth}
    >{\raggedleft\arraybackslash}p{0.04\linewidth}
    >{\raggedleft\arraybackslash}p{0.04\linewidth} 
    >{\raggedleft\arraybackslash}p{0.04\linewidth}
    >{\raggedleft\arraybackslash}p{0.04\linewidth}|
    >{\raggedright\arraybackslash}p{0.2\linewidth}
    >{\raggedleft\arraybackslash}p{0.04\linewidth}
    >{\raggedleft\arraybackslash}p{0.04\linewidth} 
    >{\raggedleft\arraybackslash}p{0.04\linewidth}
    >{\raggedleft\arraybackslash}p{0.05\linewidth}
}
\toprule
                       Dataset &    $N$ &  $N_+$ &  $f$ &    $IR$ &                    Dataset &    $N$ &  $N_+$ &  $f$ &     $IR$ \\
\midrule
                        glass1 &  214 &  76 &  9 &  1.82 &              yeast-2\_vs\_8 &  482 &  20 &  8 &  23.10 \\
                        yeast1 & 1484 & 429 &  8 &  2.46 &                   flare-F & 1066 &  43 & 11 &  23.79 \\
                      haberman &  306 &  81 &  3 &  2.78 &                  car-good & 1728 &  69 &  6 &  24.04 \\
                      vehicle3 &  846 & 212 & 18 &  2.99 &  kr-vs-k-zero-one\_vs\_draw & 2901 & 105 &  6 &  26.63 \\
                        ecoli1 &  336 &  77 &  7 &  3.36 &                    yeast4 & 1484 &  51 &  8 &  28.10 \\
                  yeast-2\_vs\_4 &  514 &  51 &  8 &  9.08 &         winequality-red-4 & 1599 &  53 & 11 &  29.17 \\
          yeast-0-3-5-9\_vs\_7-8 &  506 &  50 &  8 &  9.12 &        yeast-1-2-8-9\_vs\_7 &  947 &  30 &  8 &  30.57 \\
      yeast-0-2-5-7-9\_vs\_3-6-8 & 1004 &  99 &  8 &  9.14 &                    yeast5 & 1484 &  44 &  8 &  32.73 \\
          yeast-0-5-6-7-9\_vs\_4 &  528 &  51 &  8 &  9.35 &   kr-vs-k-three\_vs\_eleven & 2935 &  81 &  6 &  35.23 \\
              glass-0-1-6\_vs\_2 &  192 &  17 &  9 & 10.29 &    abalone-17\_vs\_7-8-9-10 & 2338 &  58 &  8 &  39.31 \\
led7digit-0-2-4-5-6-7-8-9\_vs\_1 &  443 &  37 &  7 & 10.97 &                    yeast6 & 1484 &  35 &  8 &  41.40 \\
                        glass2 &  214 &  17 &  9 & 11.59 & abalone-19\_vs\_10-11-12-13 & 1622 &  32 &  8 &  49.69 \\
                   abalone9-18 &  731 &  42 &  8 & 16.40 &            poker-8-9\_vs\_5 & 2075 &  25 & 10 &  82.00 \\
                        glass5 &  214 &   9 &  9 & 22.78 &                 abalone19 & 4174 &  32 &  8 & 129.44 \\
\bottomrule
\end{tabular*}
\end{threeparttable}
\end{table*}

\begin{algorithm}[!h]
\caption{GAT-RWOS}
\label{alg:gat_rwos}
\footnotesize
\begin{algorithmic}[1]
\Require $G$: Graph representing the dataset, $\mathbf{X}$: Feature matrix, $\mathbf{y}$: Labels, $\alpha$: Attention weight threshold, $n_s$: Number of synthetic samples, $n_w$: Number of random walk steps, $n_i$: Number of interpolations, $p$: Return probability, $q$: In-out probability
\Ensure $\mathbf{X}_\text{syn}$: Synthetic minority samples
\State Train a GAT on $G$ to obtain attention weights $\mathbf{A}$
\State Create an attention matrix $\mathbf{M}$ by aggregating attention weights
\State Identify minority nodes $V_m$ based on $\mathbf{y}$
\State Extract the minority attention matrix $\mathbf{M}_m$
\State Select strong minority connections $E_s$ based on $\alpha$
\State Calculate the number of synthetic samples $n_s$
\State $\mathbf{X}_\text{syn} \gets \emptyset$
\While{$|\mathbf{X}_\text{syn}| < n_s$}
    \State Select a pair of strongly connected minority nodes $(v_i, v_j)$ from $E_s$
    \State $P \gets \textsc{BiasedWalk}(v_i, \mathbf{M}_m, n_w, p, q)$
    \If{$|P| < 2$}
        \State \textbf{continue}
    \EndIf
    \State Calculate adaptive interpolation weights along $P$
    \For{$k \gets 1$ to $n_i$}
        \State Select consecutive nodes $(v_a, v_b)$ from $P$
        \State $\mathbf{x}_\text{new} \gets \textsc{AttInterp}(\mathbf{x}_a, \mathbf{x}_b, \text{weight}, \mathcal{C}, \mathcal{T})$
        \State $\mathbf{X}_\text{syn} \gets \mathbf{X}_\text{syn} \cup \mathbf{x}_\text{new}$
    \EndFor
\EndWhile
\State \Return $\mathbf{X}_\text{syn}$
\end{algorithmic}
\end{algorithm}

\subsection{Data Preparation}

The data preprocessing involved standardizing the target class labels so that the positive class is always the minority class across all datasets. The datasets were split into 80\% training, 10\% validation, and 10\% testing sets. The training set was used for GAT training and balancing, the validation set for parameter optimization, and the testing set for performance evaluation. All features were retained, with categorical variables label encoded and features scaled to [0, 1] to prevent scale bias.

Since our GAT algorithm requires graph inputs, tabular datasets were transformed using distance metrics (Manhattan, Euclidean, and Cosine) to generate a similarity matrix \( S \) of size \( N \times N \), where \( N \) is the number of instances. Each element \( S_{ij} \) represents the similarity between instances \( i \) and \( j \). We constructed a graph \( G \) from \( S \), using similarities as edge weights and applying a similarity threshold \( \tau \) to filter out irrelevant edges.

\subsection{Machine Learning Models}

To assess the effectiveness of the synthetic samples in balancing the training set, we employed various ML models: Logistic Regression (LR), Random Forest (RF), k-Nearest Neighbors (KNN), Extreme Gradient Boosting (XGB) and Naive Bayes (NB). Default hyperparameters from the respective libraries were used to ensure consistency and avoid manual tuning bias.

\subsection{Evaluation Metrics}\label{subsec:evaluation_metrics}

In imbalanced learning, improving minority class performance while maintaining majority class effectiveness is crucial. Following \cite{Kovacs2019}, we selected four evaluation metrics: balanced accuracy, F1-Score, ROC AUC Score, and G-Mean.
Balanced accuracy computes the average recall of each class:
\[
AC_{\text{balanced}} = \frac{1}{2} \left( \frac{TP}{TP + FN} + \frac{TN}{TN + FP} \right)
\]
The F1-Score, the harmonic mean of precision and recall, is particularly useful for imbalanced classes:
\[
F1 = 2 \times (\text{Precision} \times \text{Recall}) / (\text{Precision} + \text{Recall})
\]
The ROC AUC Score measures the area under the receiver operating characteristic curve, indicating the model's ability to discriminate between classes. The G-Mean, assessing the balance between class performances, is calculated as:
\[
G = \sqrt{\text{Precision} \times \text{Recall}}
\]
These metrics collectively provide a comprehensive evaluation of GAT-RWOS's handling of class imbalances in binary classification tasks.

\subsection{Hyperparameter Tuning}\label{subsec:tuning}

Given the complexity of our approach, hyperparameter tuning was essential. We employed Bayesian optimization using the Optuna framework \cite{akiba_optuna_2019} to efficiently navigate the parameter space. Parameters tuned included the distance metric type, similarity threshold (\( \tau \)), attention threshold (\( \theta \)), GAT parameters (number of input and output heads, hidden units, dropout rate), and biased random walk parameters ($p$, $q$, number of steps). We conducted five optimization trials, each limited to 30 minutes, to improve the $F1$ score on validation data.

\subsection{Hardware and Software Specifications}\label{subsec:specs}
The experiments were conducted on a Windows 10 system with Python 3.10, with an Intel Core i7-12700 CPU (12 cores, 20 threads), 128\,GB RAM, and an NVIDIA GeForce RTX 4090 GPU (24\,GB VRAM).

\section{Results \& Discussion}\label{sec:results}

\begin{table*}[!ht]
\caption{Performance Evaluation of Oversampled Data using GAT-RWOS and Original Data.}\label{table:results}
\scriptsize
\centering
\begin{threeparttable}
\begin{tabular*}{0.8\textwidth}{@{\extracolsep{\fill}} 
    >{\raggedright\arraybackslash}p{0.22\linewidth}
    >{\centering\arraybackslash}p{0.05\linewidth}|
    >{\centering\arraybackslash}p{0.05\linewidth}
    >{\centering\arraybackslash}p{0.05\linewidth} 
    >{\centering\arraybackslash}p{0.05\linewidth}
    >{\centering\arraybackslash}p{0.05\linewidth}|
    >{\centering\arraybackslash}p{0.05\linewidth}
    >{\centering\arraybackslash}p{0.05\linewidth}
    >{\centering\arraybackslash}p{0.05\linewidth}
    >{\centering\arraybackslash}p{0.05\linewidth} 
}
\toprule
                         &      & \multicolumn{4}{c|}{\textbf{Oversampled Data}} & \multicolumn{4}{c}{\textbf{Original Data}} \\
                       Dataset &     $IR$ & $AC_{\text{balanced}}$ &  $F1$ & $\text{AUC}$ &   $G$ & $AC_{\text{balanced}}$ &  $F1$ & $\text{AUC}$ &   $G$ \\
\midrule
                        glass1 &   1.82 &           0.866 & 0.824 & 0.866 & 0.866 &           0.830 & 0.778 & 0.830 & 0.829 \\
                        yeast1 &   2.46 &           0.771 & 0.674 & 0.771 & 0.765 &           0.680 & 0.543 & 0.680 & 0.659 \\
                      haberman &   2.78 &           0.766 & 0.632 & 0.766 & 0.766 &           0.707 & 0.571 & 0.707 & 0.676 \\
                      vehicle3 &   2.99 &           0.826 & 0.778 & 0.826 & 0.810 &           0.786 & 0.703 & 0.786 & 0.768 \\
                        ecoli1 &   3.36 &           0.904 & 0.762 & 0.904 & 0.899 &           0.774 & 0.667 & 0.774 & 0.760 \\
                  yeast-2\_vs\_4 &   9.08 &           0.889 & 0.800 & 0.889 & 0.885 &           0.789 & 0.667 & 0.789 & 0.766 \\
          yeast-0-3-5-9\_vs\_7-8 &   9.12 &           1.000 & 1.000 & 1.000 & 1.000 &           0.867 & 0.667 & 0.867 & 0.865 \\
      yeast-0-2-5-7-9\_vs\_3-6-8 &   9.14 &           0.845 & 0.778 & 0.845 & 0.832 &           0.800 & 0.750 & 0.800 & 0.775 \\
          yeast-0-5-6-7-9\_vs\_4 &   9.35 &           0.969 & 0.769 & 0.969 & 0.968 &           0.700 & 0.571 & 0.700 & 0.632 \\
              glass-0-1-6\_vs\_2 &  10.29 &           1.000 & 1.000 & 1.000 & 1.000 &           0.694 & 0.267 & 0.694 & 0.624 \\
led7digit-0-2-4-5-6-7-8-9\_vs\_1 &  10.97 &           0.863 & 0.750 & 0.863 & 0.855 &           0.863 & 0.750 & 0.863 & 0.855 \\
                        glass2 &  11.59 &           1.000 & 1.000 & 1.000 & 1.000 &           0.525 & 0.167 & 0.525 & 0.524 \\
                   abalone9-18 &  16.40 &           0.875 & 0.857 & 0.875 & 0.866 &           0.750 & 0.667 & 0.750 & 0.707 \\
                        glass5 &  22.78 &           1.000 & 1.000 & 1.000 & 1.000 &           0.976 & 0.667 & 0.976 & 0.976 \\
                  yeast-2\_vs\_8 &  23.10 &           0.750 & 0.667 & 0.750 & 0.707 &           0.617 & 0.100 & 0.617 & 0.484 \\
                       flare-F &  23.79 &           0.865 & 0.667 & 0.865 & 0.858 &           0.500 & 0.000 & 0.500 & 0.000 \\
                      car-good &  24.04 &           0.929 & 0.923 & 0.929 & 0.926 &           0.783 & 0.667 & 0.783 & 0.754 \\
      kr-vs-k-zero-one\_vs\_draw &  26.63 &           0.855 & 0.667 & 0.855 & 0.845 &           0.675 & 0.421 & 0.675 & 0.599 \\
                        yeast4 &  28.10 &           0.797 & 0.667 & 0.797 & 0.772 &           0.593 & 0.250 & 0.593 & 0.444 \\
             winequality-red-4 &  29.17 &           0.600 & 0.333 & 0.600 & 0.447 &           0.500 & 0.000 & 0.500 & 0.000 \\
            yeast-1-2-8-9\_vs\_7 &  30.57 &           0.667 & 0.500 & 0.667 & 0.577 &           0.538 & 0.066 & 0.538 & 0.276 \\
                        yeast5 &  32.73 &           0.990 & 0.727 & 0.990 & 0.990 &           0.743 & 0.500 & 0.743 & 0.702 \\
       kr-vs-k-three\_vs\_eleven &  35.23 &           0.872 & 0.750 & 0.872 & 0.863 &           0.802 & 0.526 & 0.802 & 0.782 \\
        abalone-17\_vs\_7-8-9-10 &  39.31 &           0.833 & 0.800 & 0.833 & 0.816 &           0.667 & 0.500 & 0.667 & 0.577 \\
                        yeast6 &  41.40 &           0.875 & 0.857 & 0.875 & 0.866 &           0.750 & 0.667 & 0.750 & 0.707 \\
     abalone-19\_vs\_10-11-12-13 &  49.69 &           0.660 & 0.333 & 0.660 & 0.574 &           0.500 & 0.000 & 0.500 & 0.000 \\
                poker-8-9\_vs\_5 &  82.00 &           0.667 & 0.500 & 0.667 & 0.577 &           0.667 & 0.500 & 0.667 & 0.577 \\
                     abalone19 & 129.44 &           0.664 & 0.333 & 0.664 & 0.576 &           0.500 & 0.000 & 0.500 & 0.000 \\
\midrule
\textbf{Average} & \textbf{25.62} & \textbf{0.843} & \textbf{0.727} & \textbf{0.843} & \textbf{0.818} & \textbf{0.699} & \textbf{0.451} & \textbf{0.699} & \textbf{0.583} \\
\bottomrule
\end{tabular*}
\end{threeparttable}
\end{table*}

We introduce GAT-RWOS, a novel oversampling method that combines GATs with random walk-based oversampling to address class imbalance. We evaluate its effectiveness on various imbalanced datasets, comparing performance against the original data and other state-of-the-art oversampling techniques. Additionally, we visualize how the synthetic samples generated by GAT-RWOS are distributed in the feature space for selected datasets.

Table \ref{table:results} compares GAT-RWOS performance to the original imbalanced data across $AC_{balanced}$, $F1$ score, $AUC$, and $G$ metrics, revealing the following key insights:
\begin{itemize}
    \item GAT-RWOS consistently outperforms the original data across all datasets and metrics, effectively addressing class imbalance and improving classification performance.
    
    \item The performance improvement is significant for datasets with high IRs. For example, in the \texttt{yeast5} dataset (IR: 32.73), GAT-RWOS raises $AC_{balanced}$ from 0.743 to 0.99 and $F1$ score from 0.5 to 0.727.
    
    \item Even for datasets with low IRs, like \texttt{glass1} (IR: 1.82), GAT-RWOS shows notable gains: $AC_{balanced}$ increases from 0.83 to 0.866, and $F1$ score from 0.778 to 0.824.
    
    \item The attention-guided random walk effectively generates informative synthetic samples, as evidenced by consistent improvements in $AUC$ and $G$ metrics, which measure the discriminative power of the classifier and the geometric mean of class-wise accuracies, respectively.
    
    \item In cases such as \texttt{glass5} and \texttt{yeast-0-3-5-9\_vs\_7-8}, GAT-RWOS achieves perfect scores of 1.0 across all metrics, fully resolving class imbalance and enabling accurate classification.
    
    \item Performance gains are more pronounced with higher IRs, indicating that GAT-RWOS is particularly well-suited for handling severe class imbalance scenarios.
\end{itemize}

The comparative results of GAT-RWOS against other oversampling methods are presented in Table \ref{table:comparative}. Key insights include:
\begin{itemize}
    \item GAT-RWOS consistently outperforms the other oversampling methods across various datasets with varying IR.
        
    \item On highly imbalanced datasets (e.g., \texttt{glass-0-1-6\_vs\_2} with IR 10.29, \texttt{glass2} with IR 11.59, and \texttt{glass5} with IR 22.78), GAT-RWOS achieves perfect F1-scores of 1.0, significantly surpassing other methods and demonstrating its effectiveness in extreme class imbalance scenarios.
    
    \item GAT-RWOS shows substantial improvements over the baseline RWOS method by incorporating an attention mechanism for more targeted oversampling. For instance, on \texttt{abalone9-18} (IR: 16.4), GAT-RWOS achieves an F1-score of 0.857 compared to 0.444 for RWOS.

    \item Compared to SMOTE and its variants (PF-SMOTE, SMOTE-IPF), GAT-RWOS generally achieves higher F1-scores and G-means, indicating its more effective generation of informative synthetic samples and improved class separability.

    \item GAT-RWOS also outperforms ProWSyn on most datasets. For example, on \texttt{yeast4} (IR: 28.1), GAT-RWOS achieves an F1-score of 0.667 compared to 0.588 for ProWSyn.
    
    \item The performance gains of GAT-RWOS are more pronounced on datasets with higher imbalance ratios, indicating that the attention mechanism effectively navigates complex decision boundaries in highly imbalanced scenarios.

\end{itemize}
Overall, these results demonstrate GAT-RWOS's superiority in handling class imbalance, proving the effectiveness of attention-guided random walks in generating informative synthetic samples and improving classification performance.

\begin{figure*}[!ht]
    \centering
    \includegraphics[width=0.7\linewidth]{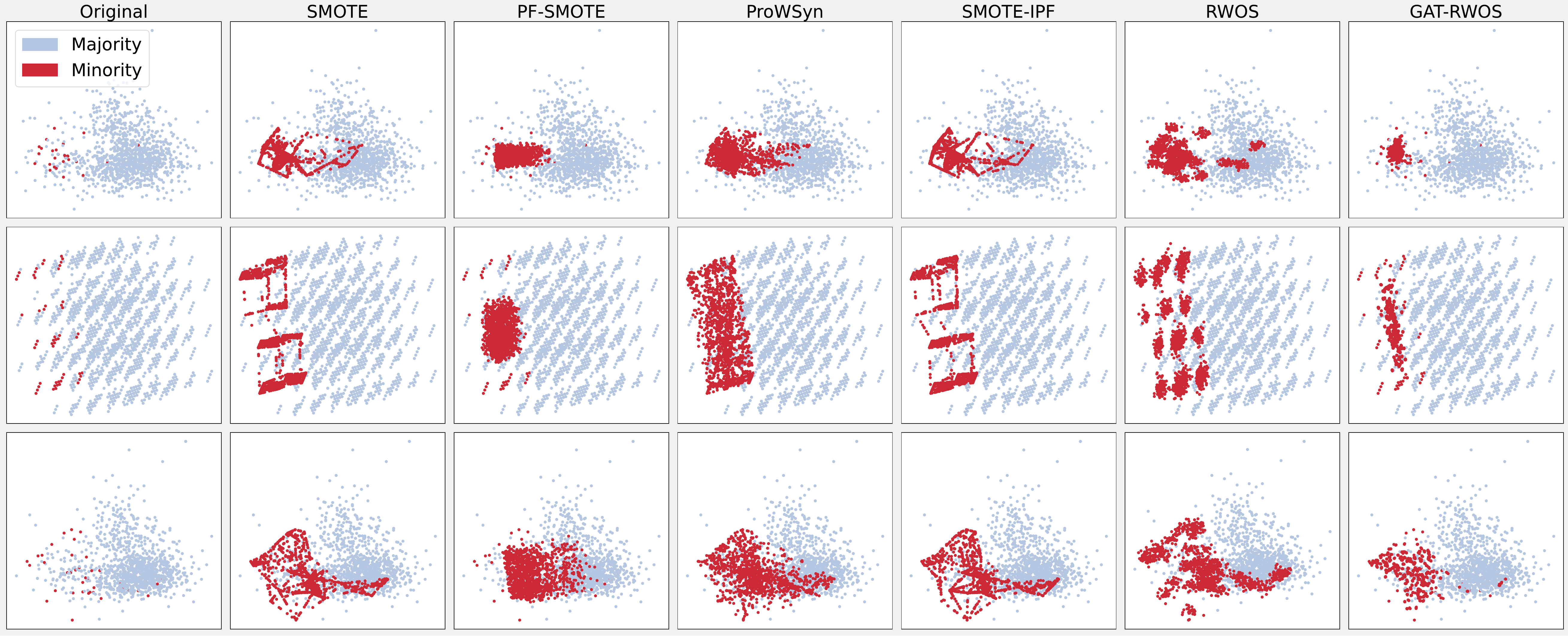}
    \caption{Placement of Synthetic Samples by Various Oversampling Methods.}
    \label{fig:sampling_examples}
\end{figure*}

Figure \ref{fig:sampling_examples} illustrates the placement of synthetic minority samples generated by various oversampling methods compared to the original imbalanced data on the datasets yeast6, car-good, and yeast4. The figure highlights the differences in how each method generates and places synthetic minority samples:

\begin{itemize}
    \item \textbf{Original Data}: Exhibits significant imbalance between majority (blue) and minority (red) classes, with minority classes underrepresented.
    \item \textbf{SMOTE}: Places new samples along paths connecting minority samples, potentially overlapping with the majority class.
    \item \textbf{PF-SMOTE}: Places samples within inner minority regions, forming a rectangular shape, without considering the majority class distribution.
    \item \textbf{ProWSyn}: Generates samples within minority boundaries without considering majority presence, possibly causing overlap.
    \item \textbf{SMOTE-IPF}: Similar to SMOTE, places samples along paths connecting minority samples, potentially overlapping with the majority class.
    \item \textbf{RWOS}: Generates samples around minority samples without considering majority distribution, possibly placing samples atop majority samples.
    \item \textbf{GAT-RWOS}: Places samples around closely situated minority samples, avoiding outliers, and minimizes overlap with the majority class.
\end{itemize}
Overall, GAT-RWOS stands out by considering local minority density and minimizing overlap with the majority class, potentially leading to better class separation and improved classification performance.

\section{Conclusion}\label{sec:conclusion}

We introduced GAT-RWOS, a novel oversampling approach that combines GATs with random walk-based oversampling. By leveraging the attention mechanism of GATs to guide the random walk and focus on the most informative neighbourhoods of minority nodes, GAT-RWOS generates high-quality synthetic samples that expand class boundaries while preserving the original data distribution.
Experiments on various imbalanced datasets demonstrate that GAT-RWOS effectively improves classification performance, consistently outperforming the original data and state-of-the-art oversampling techniques.

Despite these promising results, GAT-RWOS has limitations. Its computational complexity is higher than simpler oversampling methods due to the training of the GAT model. Additionally, its effectiveness on multi-class imbalanced problems remains to be explored.

Future work includes optimizing GAT-RWOS's computational efficiency, extending it to handle multi-class imbalance. Combining GAT-RWOS with instance selection methods to remove redundant instances may further enhance the quality of synthetic samples. Applying GAT-RWOS to real-world applications like fraud detection, medical diagnosis, and anomaly detection could validate its practical utility.


\begin{table*}[!ht]
\caption{Comparison of Various Oversamplers based on $F1$ and $G$ Metrics.}\label{table:comparative}
\centering
\scriptsize
\begin{threeparttable}
\begin{tabular*}{0.85\textwidth}{@{\extracolsep{\fill}} 
    >{\raggedright\arraybackslash}p{0.18\linewidth}
    >{\centering\arraybackslash}p{0.05\linewidth}
    >{\centering\arraybackslash}p{0.05\linewidth}
    >{\centering\arraybackslash}p{0.08\linewidth}
    >{\centering\arraybackslash}p{0.08\linewidth} 
    >{\centering\arraybackslash}p{0.08\linewidth}
    >{\centering\arraybackslash}p{0.09\linewidth}
    >{\centering\arraybackslash}p{0.08\linewidth}
    >{\centering\arraybackslash}p{0.08\linewidth}
}
\toprule
 Dataset & $IR$ & Metric   &  SMOTE &  PF-SMOTE &  ProWSyn &  SMOTE-IPF &   RWOS &  GAT-RWOS \\
\midrule
\multirow{2}{*}{glass1} & \multirow{2}{*}{1.82} & $F1$ &  0.778 &     \textbf{0.824} &    0.706 &      0.778 &  0.778 &     \textbf{0.824} \\
          &        & $G$ &  0.829 &     \textbf{0.866} &    0.768 &      0.829 &  0.829 &     \textbf{0.866} \\
\midrule

\multirow{2}{*}{yeast1} & \multirow{2}{*}{2.46} & $F1$ &  0.636 &     0.561 &    0.629 &      0.622 &  0.625 &     \textbf{0.674} \\
          &        & $G$ &  0.739 &     0.674 &    0.735 &      0.731 &  0.718 &     \textbf{0.765} \\
\midrule

\multirow{2}{*}{haberman} & \multirow{2}{*}{2.78} & $F1$ &  0.526 &     0.571 &    0.571 &      0.500 &  0.500 &     \textbf{0.632} \\
          &        & $G$ &  0.680 &     0.676 &    0.676 &      0.643 &  0.599 &     \textbf{0.766} \\
\midrule

\multirow{2}{*}{vehicle3} & \multirow{2}{*}{2.99} & $F1$ &  0.634 &     0.632 &    0.571 &      0.737 &  0.703 &     \textbf{0.778} \\
          &        & $G$ &  0.743 &     0.726 &    0.701 &      0.797 &  0.768 &     \textbf{0.810} \\
\midrule

\multirow{2}{*}{ecoli1} & \multirow{2}{*}{3.36} & $F1$ &  0.700 &     0.700 &    0.700 &      \textbf{0.762} &  0.737 &     \textbf{0.762} \\
          &        & $G$ &  0.841 &     0.841 &    0.841 &      \textbf{0.899} &  0.860 &     \textbf{0.899} \\
\midrule

\multirow{2}{*}{yeast-2\_vs\_4} & \multirow{2}{*}{9.08} & $F1$ &  0.571 &     0.727 &    0.600 &      0.615 &  0.727 &     \textbf{0.800} \\
          &        & $G$ &  0.846 &     0.875 &    0.758 &      0.856 &  0.875 &     \textbf{0.885} \\
\midrule

\multirow{2}{*}{yeast-0-3-5-9\_vs\_7-8} & \multirow{2}{*}{9.12} & $F1$ &  0.667 &     0.500 &    0.727 &      0.800 &  0.667 &     \textbf{1.000} \\
          &        & $G$ &  0.865 &     0.824 &    0.875 &      0.885 &  0.865 &     \textbf{1.000} \\
\midrule

\multirow{2}{*}{yeast-0-2-5-7-9\_vs\_3-6-8} & \multirow{2}{*}{9.14} & $F1$ &  0.700 &     0.750 &    \textbf{0.778} &      0.706 &  0.737 &     \textbf{0.778} \\
          &        & $G$ &  0.823 &     0.775 &    \textbf{0.832} &      0.770 &  0.827 &     \textbf{0.832} \\
\midrule

\multirow{2}{*}{yeast-0-5-6-7-9\_vs\_4} & \multirow{2}{*}{9.35} & $F1$ &  0.667 &     0.714 &    0.750 &      0.615 &  0.667 &     \textbf{0.769} \\
          &        & $G$ &  0.946 &     0.957 &    0.775 &      0.856 &  0.946 &     \textbf{0.968} \\
\midrule

\multirow{2}{*}{glass-0-1-6\_vs\_2} & \multirow{2}{*}{10.29} & $F1$ &  0.444 &     0.500 &    0.667 &      0.444 &  0.500 &     \textbf{1.000} \\
          &        & $G$ &  0.850 &     0.687 &    0.707 &      0.850 &  0.687 &     \textbf{1.000} \\
\midrule

\multirow{2}{*}{led7digit-0-2-4-5-6-7-8-9\_vs\_1} & \multirow{2}{*}{10.97} & $F1$ &  0.667 &     \textbf{0.750} &    0.667 &      \textbf{0.750} &  \textbf{0.750} &     \textbf{0.750} \\
          &        & $G$ &  0.845 &     \textbf{0.855} &    0.845 &      \textbf{0.855} &  \textbf{0.855} &     \textbf{0.855} \\
\midrule

\multirow{2}{*}{glass2} & \multirow{2}{*}{11.59} & $F1$ &  0.667 &     0.500 &    0.667 &      0.667 &  0.308 &     \textbf{1.000} \\
          &        & $G$ &  0.707 &     0.894 &    0.707 &      0.707 &  0.742 &     \textbf{1.000} \\
\midrule

\multirow{2}{*}{abalone9-18} & \multirow{2}{*}{16.40} & $F1$ &  0.545 &     0.400 &    0.444 &      0.429 &  0.444 &     \textbf{0.857} \\
          &        & $G$ &  0.841 &     0.687 &    \textbf{0.926} &      0.822 &  0.692 &     0.866 \\
\midrule

\multirow{2}{*}{glass5} & \multirow{2}{*}{22.78} & $F1$ &  0.667 &     0.667 &    0.667 &      0.667 &  0.667 &     \textbf{1.000} \\
          &        & $G$ &  0.976 &     0.976 &    0.976 &      0.976 &  0.976 &     \textbf{1.000} \\
\midrule

\multirow{2}{*}{yeast-2\_vs\_8} & \multirow{2}{*}{23.10} & $F1$ &  0.098 &     0.182 &    0.114 &      0.111 &  0.089 &     \textbf{0.667} \\
          &        & $G$ &  0.461 &     \textbf{0.786} &    0.583 &      0.565 &  0.357 &     0.707 \\
\midrule

\multirow{2}{*}{flare-F} & \multirow{2}{*}{23.79} & $F1$ &  0.211 &     0.571 &    0.167 &      0.211 &  0.211 &     \textbf{0.667} \\
          &        & $G$ &  0.842 &     0.704 &    0.734 &      0.842 &  0.842 &     \textbf{0.858} \\
\midrule

\multirow{2}{*}{car-good} & \multirow{2}{*}{24.04} & $F1$ &  0.857 &     0.769 &    0.833 &      0.857 &  0.778 &     \textbf{0.923} \\
          &        & $G$ &  0.923 &     0.843 &    0.845 &      0.923 &  \textbf{0.988} &     0.926 \\
\midrule

\multirow{2}{*}{kr-vs-k-zero-one\_vs\_draw} & \multirow{2}{*}{26.63} & $F1$ &  0.500 &     0.500 &    0.632 &      0.478 &  0.500 &     \textbf{0.667} \\
          &        & $G$ &  0.602 &     0.602 &    0.736 &      \textbf{0.956} &  0.602 &     0.845 \\
\midrule

\multirow{2}{*}{yeast4} & \multirow{2}{*}{28.10} & $F1$ &  0.400 &     0.273 &    0.588 &      0.462 &  0.294 &     \textbf{0.667} \\
          &        & $G$ &  0.626 &     0.736 &    \textbf{0.975} &      0.761 &  0.913 &     0.772 \\
\midrule

\multirow{2}{*}{winequality-red-4} & \multirow{2}{*}{29.17} & $F1$ &  0.250 &     0.062 &    0.143 &      0.250 &  0.222 &     \textbf{0.333} \\
          &        & $G$ &  0.444 &     0.408 &    0.436 &      0.444 &  0.443 &     \textbf{0.447} \\
\midrule

\multirow{2}{*}{yeast-1-2-8-9\_vs\_7} & \multirow{2}{*}{30.57} & $F1$ &  0.222 &     0.182 &    0.167 &      0.400 &  0.286 &     \textbf{0.500} \\
          &        & $G$ &  \textbf{0.878} &     0.737 &    0.727 &      0.574 &  0.776 &     0.577 \\
\midrule

\multirow{2}{*}{yeast5} & \multirow{2}{*}{32.73} & $F1$ &  0.471 &     0.471 &    0.545 &      0.471 &  0.600 &     \textbf{0.727} \\
          &        & $G$ &  0.968 &     0.968 &    0.854 &      0.968 &  0.857 &     \textbf{0.990} \\
\midrule

\multirow{2}{*}{kr-vs-k-three\_vs\_eleven} & \multirow{2}{*}{35.23} & $F1$ &  0.526 &     0.667 &    0.667 &      0.625 &  0.667 &     \textbf{0.750} \\
          &        & $G$ &  0.782 &     0.788 &    0.788 &      0.786 &  0.788 &     \textbf{0.863} \\
\midrule

\multirow{2}{*}{abalone-17\_vs\_7-8-9-10} & \multirow{2}{*}{39.31} & $F1$ &  0.429 &     0.667 &    0.571 &      0.500 &  0.500 &     \textbf{0.800} \\
          &        & $G$ &  0.699 &     0.707 &    0.809 &      0.702 &  0.702 &     \textbf{0.816} \\
\midrule

\multirow{2}{*}{yeast6} & \multirow{2}{*}{41.40} & $F1$ &  0.667 &     0.750 &    0.667 &      0.667 &  0.750 &     \textbf{0.857} \\
          &        & $G$ &  0.860 &     0.863 &    0.860 &      0.860 &  0.863 &     \textbf{0.866} \\
\midrule

\multirow{2}{*}{abalone-19\_vs\_10-11-12-13} & \multirow{2}{*}{49.69} & $F1$ &  0.190 &     0.130 &    0.143 &      0.190 &  0.140 &     \textbf{0.333} \\
          &        & $G$ &  0.775 &     0.866 &    0.756 &      0.775 &  \textbf{0.877} &     0.574 \\
\midrule

\multirow{2}{*}{poker-8-9\_vs\_5} & \multirow{2}{*}{82.00} & $F1$ &  0.062 &     0.333 &    0.065 &      0.062 &  \textbf{0.500} &     \textbf{0.500} \\
          &        & $G$ &  0.749 &     0.575 &    \textbf{0.759} &      0.749 &  0.577 &     0.577 \\
\midrule

\multirow{2}{*}{abalone19} & \multirow{2}{*}{129.44} & $F1$ &  0.065 &     0.044 &    0.125 &      0.250 &  0.061 &     \textbf{0.333} \\
          &        & $G$ &  \textbf{0.889} &     0.827 &    0.569 &      0.575 &  0.881 &     0.576 \\
\bottomrule
\end{tabular*}
\end{threeparttable}
\end{table*}

\bibliographystyle{ieeetr}

\end{document}